# Learned Focused Plenoptic Image Compression with Microimage Preprocessing and Global Attention

Kedeng Tong, Xin Jin\*, *Senior Member, IEEE*, Yuqing Yang, Chen Wang, Jinshi Kang, Fan Jiang

*Abstract*—Focused plenoptic cameras can record spatial and angular information of the light field (LF) simultaneously with higher spatial resolution relative to traditional plenoptic cameras, which facilitate various applications in computer vision. However, the existing plenoptic image compression methods present ineffectiveness to the captured images due to the complex micro-textures generated by the microlens relay imaging and long-distance correlations among the microimages. In this paper, a lossy end-to-end learning architecture is proposed to compress the focused plenoptic images efficiently. First, a data preprocessing scheme is designed according to the imaging principle to remove the sub-aperture image ineffective pixels in the recorded light field and align the microimages to the rectangular grid. Then, the global attention module with large receptive field is proposed to capture the global correlation among the feature maps using pixel-wise vector attention computed in the resampling process. Also, a new image dataset consisting of 1910 focused plenoptic images with content and depth diversity is built to benefit training and testing. Extensive experimental evaluations demonstrate the effectiveness of the proposed approach. It outperforms intra coding of HEVC and VVC by an average of 62.57% and 51.67% bitrate reduction on the 20 preprocessed focused plenoptic images, respectively. Also, it achieves 18.73% bitrate saving and generates perceptually pleasant reconstructions compared to the state-of-the-art end-to-end image compression methods, which benefits the applications of focused plenoptic cameras greatly. The dataset and code are publicly available at https://github.com/VincentChandelier/GACN.

## I. Introduction

THE focused plenoptic camera [1] is the second generation of plenoptic cameras, which can record the four-dimensional (4D) light field (LF) with higher spatial resolution. Comparing to the traditional plenoptic cameras [2], so-called plenoptic 1.0 cameras, which place the microlens array (MLA) at the image plane of the main lens, as shown in Fig. 1(a), to record the rays coming from each object point by the pixels beneath each microlens, as shown in Fig. 2(a), focused plenoptic cameras, as shown in Fig. 1(b), build a relay imaging architecture by putting the MLA at a distance of *a* behind the image plane of the main lens to re-image the main lens image at different angles. Each microlens in it works like a virtual camera to capture a portion of the main lens image from a perspective to form a microimage, e. g. the 69 × 69 circular regions shown in Fig. 2(b) that are captured by Tsinghua Single-focused Plenoptic Camera (TSPC) [3], and neighboring microimages record the object points from different perspectives to preserve the angular information. So, it achieves $a/b$ times higher LF spatial resolution compared to the traditional plenoptic cameras [1], where *b* is the distance from MLA to the sensor in Fig. 1(b). Combining with its portability, the focused plenoptic camera becomes one of the promising acquisition devices in 3D reconstruction, industrial inspection, and 6-degree-of-freedom virtual reality [4], which draws great attention both from academy and industry to compress the images captured by them efficiently. ISO/IEC JTC 1/SC 29/WG 04 established an ad hoc group of lenslet video coding [5, 6] to study the compression techniques for it. JPEG PLENO [7] also starts to analyze the data for standardization. However, efficient compression is still challenging because of the distinct pixel distribution features in the focused plenoptic images.

Comparing the features in the focused plenoptic image and those in the traditional plenoptic image for the same scene in Fig. 2, significant distribution differences can be found. As shown in the magnifications of the wheel, each microimage in the focused plenoptic image (Fig. 2(b)) is much larger than that in the traditional plenoptic image (Fig. 2(a)) because of the relay imaging architecture and each of them images the wheel from a different perspective, which makes the textures inside the microimage more complex than those in traditional plenoptic microimage. Since the views of the wheel are recorded by the neighboring microimages, as shown in Fig. 2(b), and the size of the microimage is big, the neighboring microimages present long-distance spatial correlations. These differences make hybrid coding tools like local prediction [8-16] ineffective in exploiting the spatial correlations while preserving the details simultaneously. Although some image-principle-guided compression methods [17-20] were proposed to generate the spatial predictor by block-wise zooming and interpolation, blocking artifacts may be introduced to the regions with detailed textures, which limits the compression performance. The sub-aperture/pseudo-video-based coding techniques [21-33] are not universal to focused plenoptic images because of the irreversibility of converting sub-aperture images back to the focused plenoptic images [1, 17]. And, learning-based image compression techniques [34-37] present ineffectiveness because

The work was supported in part by National Natural Science Foundation of China under Grant No. 62131011, and Shenzhen Project under Grant No. JCYJ20200109142810146.

The authors are with the Shenzhen International Graduate School, Tsinghua University, Shenzhen 518055, China (e-mail:tkd20@mails.tsinghua.edu.cn; jin.xin@sz.tsinghua.edu.cn; yangyq22@mails.tsinghua.edu.cn; wang-che20@mails.tsinghua.edu.cn; kjs20@mails.tsinghua.edu.cn; jf19@mails.tsinghua.edu.cn).

of the limited receptive field and the incompatibility between the network trained on the continuous textures in the square patches of the natural images and the textural discontinuities in hexagonally arranged microimages.

So, in this paper, a lossy end-to-end learning architecture is proposed to compress the focused plenoptic images efficiently. First, based on the observations that inter-microimage pixels, boundary incomplete microimages, and vignetting pixels in the microimages are ineffective in light field applications, like refocusing, rendering, etc., a sub-aperture image lossless preprocessing scheme is proposed to reshape the sub-aperture image effective pixels in each microimage and align the cropped microimages to the rectangular grid to be compatible with patch-based training and to reduce the pixel redundancy. Then, a global attention module with large receptive field is proposed to capture the long-distance correlations among the feature maps in the resampling process to improve compression efficiency. The query derived from the down-sampling/up-sampling feature map computes pixel-wise vector attention with the key-value pairs derived from the unsampled feature map, which makes the output of attention computing with a skip connection can draw global dependencies from the inputs. A dataset with 1910 focused plenoptic images captured by TSPC is proposed to benefit focused plenoptic image related learning applications. Extensive experimental results demonstrate that the proposed work can achieve an average of 62.57% and 51.67% bitrate reduction compared to HEVC and VVC intra coding, respectively, on 20 preprocessed plenoptic images. It also saves 18.73% bitrate relative to the state-of-the-art (SOTA) learned image compression models. It achieves SOTA performance on both peak signal-to-noise ratio (PSNR) and multi-scale-structural similarity index (MS-SSIM) in comparison with compressing the rendered sub-aperture images using HEVC inter coding. To the best of our knowledge, it is the first work that compresses focused plenoptic images directly while achieving competitive performance to sub-aperture image compression.

The remainder of this paper is organized as follows. Section II presents the related works on plenoptic image compression and learned image compression. Section III introduces the analysis of focused plenoptic images and the proposed data preprocessing scheme. Section IV introduces the proposed focused plenoptic image dataset. The learned image compression framework and the details are described in Section V. Experimental results and analysis are provided in Section VI followed by a conclusion drawn in Section VII.

## II. RELATED WORKS

In this section, plenoptic image compression methods and learned image compression approaches are introduced.

### A. Plenoptic Image Compression

The existing plenoptic image compression approaches can be mainly categorized into two groups: compressing the sub-aperture images rendered from the plenoptic image and compressing plenoptic images directly. Methods of the first category consider reordering or sparse sampling the sub-aperture images to be the pseudo-videos and utilize the inter-coding tools to exploit the correlations among the views for higher compression efficiency [21-33]. Reordering all sub-aperture images into a pseudo-video sequence with zigzag [26], raster [28], spiral [27], and hierarchical [25, 31] scan orders have been exploited by the off-the-shelf codec to reduce the spatial and angular redundancy. Alves et al. [38] employed the 4D block partition, 4D transform, and 4D coefficient to encode the 4D sub-aperture images and proposed 4D-native hexadeca-tree-oriented bit-plane clustering to improve rate-distortion performance. Bakir et al. [21] proposed to encode the sparsely selected views and reconstruct the unselected views by the linear approximation at the decoder side. Huang et al. [22] transmitted the disparity maps additionally to geometrically predict the unselected views from the decoded views. Liu et al. [33] proposed three elaborate subnetworks to reconstruct and refine dense sub-aperture images at the decoder side. Although sub-aperture images can be extracted both from traditional and focused plenoptic images, it is irreversible to convert sub-aperture images back to the focused plenoptic image due to the relay imaging principle where patches of each microimages are interpolated and tiled to obtain the rendered views [39, 40]. Thus, the sub-aperture image compression approaches that exploit the spatial-angular redundancy among the views limit the applications of the decoded focused plenoptic data [41].

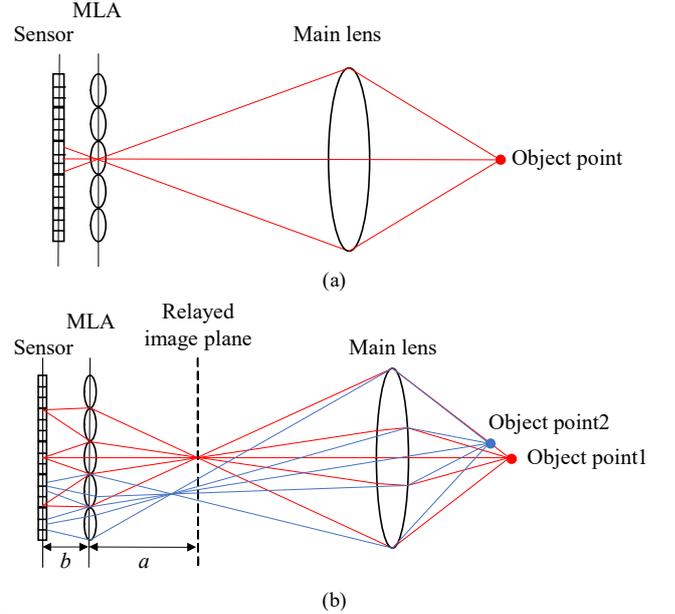

Fig. 1. Imaging principle of: (a) traditional plenoptic cameras; (b) focused plenoptic cameras. Red lines and blue lines represent the light rays coming from a focused object point and an unfocused object point, respectively.

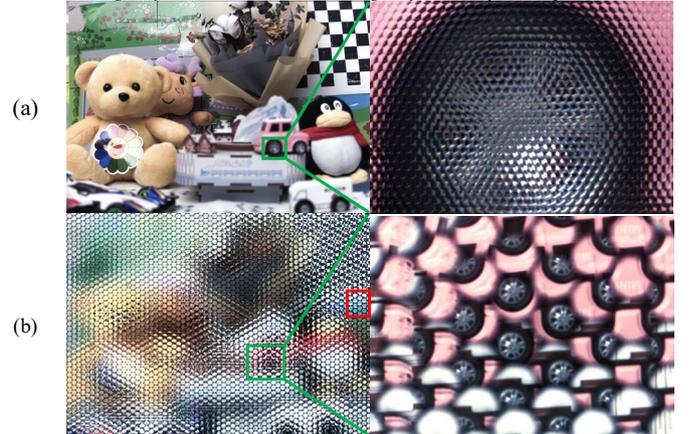

Fig. 2. Plenoptic images of a same scene captured by: (a) the traditional plenoptic camera; and (b) the focused plenoptic camera.

The second group of plenoptic image compression methods directly encodes the images to reduce the spatial redundancy of the repetitive patterns among the adjacent macropixels/ microimages [8-16, 42-44]. Locally linear embedding (LLE) [8] regards the current coding macropixels as the linear combination of the neighboring macropixels. The bi-directional self-similarity methods [9, 10] introduce two predictors derived from the adjacent macropixels for the current block and achieve impressive bitrate saving relative to HEVC. A two-stage high-order intra-prediction method [11] predicts each coding unit through the geometric transformation within the decoded region. The regression-based prediction method [12] proposes to classify the coding units into three categories based on the content property and design content-based weighted reference blocks to improve coding efficiency. Disparity correlation [13] and content depth [42] reveal that the depth of scenes is beneficial to find the imaging response blocks. Based on reshaping and aligning the macropixels with the rectangular grid to generate the plenoptic image with higher spatial correlation, the regularized macropixels compression approaches [14-16] improve the prediction efficiency by designing a single block or weighted multiblock prediction. The decorrelation network [44] utilizes the stride convolution and dilation convolution as specific feature extractors to decouple the spatial-angular information and the decoupled feature maps are compressed jointly to improve the compression efficiency of preprocessed traditional plenoptic images. However, these compression methods based on the short-distance spatial-angular correlation of macropixels are inevitably inapplicable and show ineffective compression performance for focused plenoptic images due to the long-distance spatial correlation and low internal redundancy in microimages.

Due to the different intensity distribution between traditional and focused plenoptic images, Li *et al.* [45] proposed a displacement intra-prediction scheme, which utilizes two reference blocks from the top and left microimages to predict the current coding unit. Image-correlated intra prediction (IIP) uses the matching blocks from the neighboring microimages, guided by the imaging principle of focused objects, as the weighted prediction for the current prediction unit (PU) to improve the compression efficiency [18]. Zoomed imaging-principle-guided prediction [17] determines the positions and scales of the prediction candidates according to the focusing status and the focal length of the microlens to improve the coding efficiency on multi-focused plenoptic images. Besides, the intra-block-copy mode of screen content coding extension (SCC) [46] shows efficiency in plenoptic image coding [18] due to the similar patterns among the adjacent microimages. However, artifacts may be introduced to the complex texture regions, which limits the compression efficiency subjectively and objectively.

*B. Learned Image Compression*

In recent years, a great surge of learned image compression approaches [34-37, 47-49] have been proposed and achieved an impressive trade-off between the bitrate and distortion compared to compression standards including JPEG [50], JPEG2000 [51], HEVC [52] and VVC [53]. Ballé *et al.* [34] proposed a simple end-to-end network to compress the factorized latent representation using differentiable quantization. Ballé *et al.* [35] proposed a hyperprior network to model the distribution of latent representation. Minnen *et al.* [37] explored the context regression model as an auxiliary model to capture the spatial redundancy in the latent space. Mixture Gaussian distribution is explored in [36, 49, 54] to estimate the accurate distribution of latent representation to improve the compression efficiency. Cheng *et al.* [36] proposed a local attention module to generate the compact latent representation to enhance the performance. Zou *et al.* [47] designed the deep window-based attention to capture the local and global redundancy to reduce the bit consumption but reconstruct high-quality images. However, limited by receptive field and distinctive data redundancy, they are still inefficient in exploring the long-distance spatial correlations among the microimages and preserving the textures in the focused plenoptic images simultaneously.

III. FOCUSED PLENOPTIC IMAGE ANALYSIS AND PROPOSED DATA PREPROCESSING SCHEME

In this section, we propose the data preprocessing method based on pixel effectiveness analysis using geometric ray tracing.

*A. Pixel Effectiveness Analysis*

Because of the relay imaging process, the pixels under each microlens of the focused plenoptic camera record a portion of the scene from a perspective, which results in the pixel effectiveness in rendering the sub-aperture images at different viewpoints differing from those in traditional plenoptic image. To analyze the pixel effectiveness in sub-aperture image rendering, TSPC designed by us, which has been adopted by ISO/IEC [55] as one of the representative focused plenoptic image acquisition devices, is selected to facilitate the analysis. As illustrated in Fig. 1(b), TSPC inserts the MLA into the light path at a distance *a* behind the image plane of the main lens and distance *b* in front of the sensor, respectively, where *a* and *b* satisfy the Gaussian equation with the focal length *f* of the MLA. Using the magnified portion in the captured raw image as an instance in Fig. 3(a), we propose to classify the pixels into four categories according to their effectiveness in optical responses:

**Inter-microimage Pixels**: To obtain the highest fill factor, circular microlenses are arranged hexagonally. The *f-number* of the main lens and the microlens is matched and the pixels under each microlens record the light rays refracting from the single microlens without cross-talk from other microlenses. Thus, the pixels located among the non-overlapping microimages, so-called inter-microimage pixels, are dark and are useless in reconstructing the recorded light field [56, 57], as shown in Fig. 3(a). These pixels are marked in black in Fig. 3(c), which takes up 9.3% of the total pixels and can be ignored in compression and transmission without loss of the spatial and angular information of the light field.

**Boundary Incomplete Microimages**: Since the size of the image sensor is generally not the integer multiples of the microlens, several incomplete microlens may exist at the boundary of the plenoptic image, under which boundary images a point away from the optical axis, the beam that can reach its image plane gradually becomes narrow due to the existence of a limited aperture, resulting in a gradual darkening effect around the periphery of the microimage, which generate so-called vignetting pixels as shown in Fig. 3(a). Although the white-image-based compensation methods [1, 40, 56] were proposed to mitigate this problem, the vignetting pixels marked in light green in Fig. 3(c) still suffer from non-negligible lens shading in the devignetted plenoptic image, as shown in Fig. 3(b), and are ineffective in sub-aperture image applications [58].

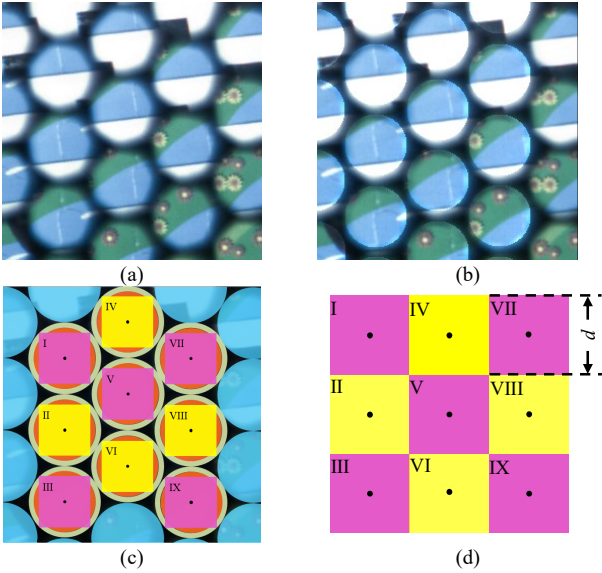

Fig. 3. Preprocessing for the 256 × 256 patch of the raw plenoptic image marked in the red rectangle in Fig. 2(b) as a toy example: (a) raw plenoptic image patch; (b) devignetted patch (a); (c) pixel classification of (b), in which inter-microimage pixels, boundary incomplete microimages, and vignetting pixels are marked in black, blue, and light green, respectively. The circular region denoted in orange contains light field effective pixels. The magenta and yellow areas covering sub-aperture image effective pixels are the pixels picked by our data preprocessing scheme; and (d) the plenoptic image patch after preprocessing, in which the magenta and yellow blocks correspond to those in (c) to show the relative position.

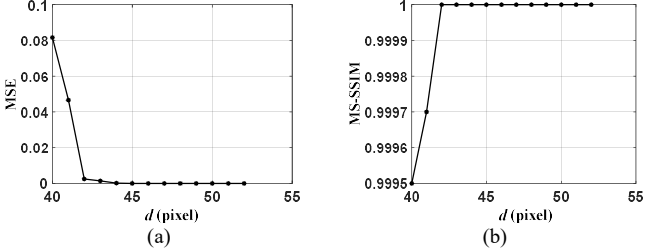

Fig. 4. Average distortion among 5 × 5 sub-aperture images with cutting size $d$. (a) Distortion measured with MSE. (b) Distortion measured with MS-SSIM.

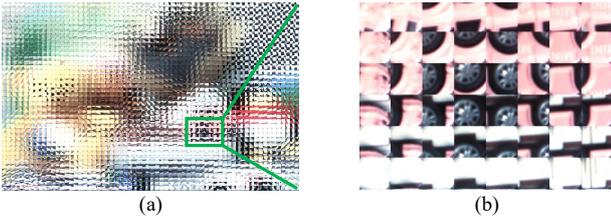

Fig. 5. Preprocessed plenoptic image and magnification. (a) Preprocessed plenoptic image. (b) Magnification of (a) in the green rectangle.

*Sub-aperture Image Effective Pixels*: Since the optical design and matching of *f-number* result in the circular exit pupil of the main lens and the microlens, the effective pixels area (EPA) underneath each complete microimages, where the light rays can reach the sensor with high energy and uniform light intensity distribution, record the reconstructive light field and are utilized for extracting the sub-aperture images from a single plenoptic image, denoted in orange in Fig 3 (c). Generally, the existing sub-aperture image conversion methods [1, 39, 40] select an inside square patch of the EPA of each microimage and tile the selected interpolated patches together to render a specific perspective. The size of selected patches depends on the depths of objects but does not exceed the size of the inscribed maximum square of EPA. In other words, sub-aperture image effective pixels located as in the inscribed maximum square of EPA are genuinely beneficial for sub-aperture image rendering.

According to the pixel effectiveness analysis, removing *Inter-microimage Pixels*, *Boundary Incomplete Microimages* and *Vignetting Pixels* will not influence the performance of sub-aperture image rendering that is mainly used in LF applications. But removing them can reduce the data amount greatly, which is beneficial to compression.

Additionally, it is found that microimages are arranged hexagonally in the plenoptic image according to the MLA layout. If we partition the plenoptic image into rectangular patches, like that generally used in the training/encoding process of end-to-end image compression [36, 37, 43, 49], massive incomplete microimages in the patches will introduce irregular intensity discontinuities, which directly harms the effectiveness of training and the feature extraction and results in low compression efficiency of transferring the existing the networks to the plenoptic images.

So, we propose a sub-aperture image lossless data preprocessing scheme by discarding the sub-aperture image ineffective pixels and reshaping the plenoptic image, which reduces data amount highly, benefits patch-based training, and does not sacrifice the LF application capabilities.

*B. Proposed Data Preprocessing Scheme*

The proposed data preprocessing scheme consists of two steps: *Microimage Cropping* and *Microimage Aligning*.

*Microimage Cropping* is to pick out sub-aperture image effective pixels in each microimages as the basic unit of preprocessed plenoptic images. Considering the imaging principle of focused plenoptic cameras and existing rendering methods, microimage cutting should guarantee the central pixel of each microimages is still the center of cropped microimage. Thus, taking the central pixels as the fixed reference points of each microimage, the square area is cropped with the cutting incomplete microimages are generated. Also, these regions generally record the light from the boundary of the main lens, which present low imaging quality. Like the regions marked in blue in Fig. 3(c), low imaging quality and insufficient spatial-angular information can be observed in Fig. 3(b) correspondingly, which cannot be used in subaperture image rendering. Taking the plenoptic images captured by TSPC as an example, the pixels of boundary incomplete microimages take about 8.3% of total pixels, which can be discarded in sub-aperture image rendering [39].

*Vignetting Pixels*: Since each microlens acts as a thin lens, as it size $d$, as the yellow and magenta areas shown in Fig. 3(c). From the analysis of the above sub-aperture image effective pixels, $d$ is given by:

$$d \geq d_{min} = \sqrt{2}mR, \quad (1)$$

where $d_{min}$, $m$ and $R$ are the length of minimum inscribed square, EPA radius coefficient, and the radius of microlens, respectively. Therefore, each complete microimage are processed into the $d \times d$ cropped microimages.

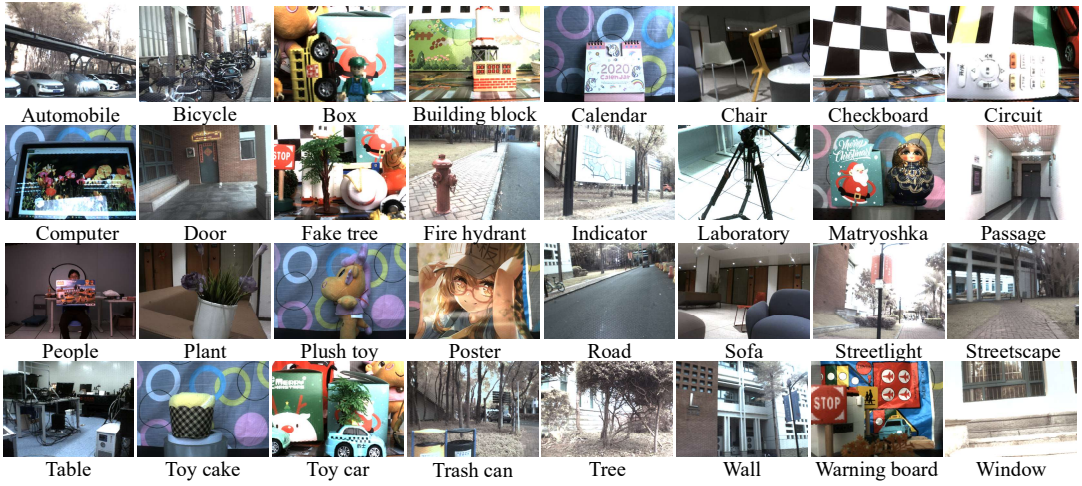

**Fig. 6.** Snapshot of the categories in the dataset. Each category is presented by the central view of a plenoptic image.

*Microimage Aligning* is to move the cropped microimages of the even column vertically to align them to the rectangular grid. The translation distance $d_{td}$ is given by:

$$d_{td} = d/2. \quad (2)$$

After *Microimage Cropping* and *Aligning*, the preprocessed plenoptic image in Fig. 3(c) becomes Fig. 3(d).

To demonstrate that the preprocessed data will not influence the sub-aperture image rendering performance, an analysis of the first frame in MPEG standard testing sequence "Cars" [59] is conducted by calculating the average distortion among the 5 × 5 sub-aperture images rendered from the original and those rendered from the preprocessed plenoptic image using the microimage stitching rendering tool [40]. $m$ equals 0.8, white image compensation [60] is applied and $R$ is set to 35 pixels according to the specification of microlens in TSPC. Mean square error (MSE) and MS-SSIM are used to measure the average distortion of sub-aperture images. The distortion statistics are retrieved and demonstrated in Fig. 4. The results reveal that when $d = 44 > d_{min} = 39.59$, the preprocessed plenoptic image is completely sub-aperture image lossless compared to the original plenoptic images.

The processed plenoptic images can be partitioned into rectangular patches to meet the needs of the deep compression network training and inference. Considering that the downsampling ratio used in the existing learned image compression is generally an integer multiple of 16 [36], the cutting size $d$ is set as 48 pixels for TSPC plenoptic images in the preprocessing scheme. Therefore, the preprocessed image of Fig. 2(b) is shown in Fig. 5.

TABLE I
SPECIFICATION OF TSINGHUA SINGLE-FOCUSED PLENOPTIC CAMERA

| | |
|---|---|
| Sensor | beA4000-62km (Basler) |
| SDK | Basler Pylon |
| Main lens | 20 mm, f/2.8 aperture |
| Microlens focal length and f-number | 4/3 mm, f/3.49 |
| Number of microlens | 69 × 51, hexagonally packed |
| Number of complete microlens | 66 × 42 |
| Microlens diameter | 381.6 um |
| Number of pixels beneath each microlens | 69×69 pixels |
| Image sensor size | 1/1.75 inch |
| Resolution of lenslet image after calibration as the original image | 4080 (H) × 3068 (V) 1.2 million pixels |
| Bit depth | 8 bit |
| Color space | RGB |

## IV. PROPOSED FOCUSED PLENOPTIC IMAGE DATASET

To train a robust model, we propose a focused plenoptic image dataset "FPI2k"[1] captured by TSPC. The specification of TSPC is listed in Table I.

To provide the data with content diversity, we capture real scenes indoor and outdoor with object depth variations. From a single plenoptic image, 5 × 5 sub-aperture images can be generated with much larger disparities one from the other. 1910 focused plenoptic images are captured and manually annotated to 32 categories based on their contents, as shown in Fig. 6. Each category selects the central view of a plenoptic image as an example to exhibit the properties of the dataset. The images taken indoors provide a large disparity among the views while objects in various depths are provided outdoors. A variety of contents and textures that do not exist in the existing datasets are included in our dataset, which greatly benefit robust algorithm training.

To the best of our knowledge, it is the first focused plenoptic image dataset with nearly 2000 images in the real circumstances.

---

[1] https://pan.baidu.com/s/1mVFOklrm2Pf_wSyn4GAPEg?pwd=u15n

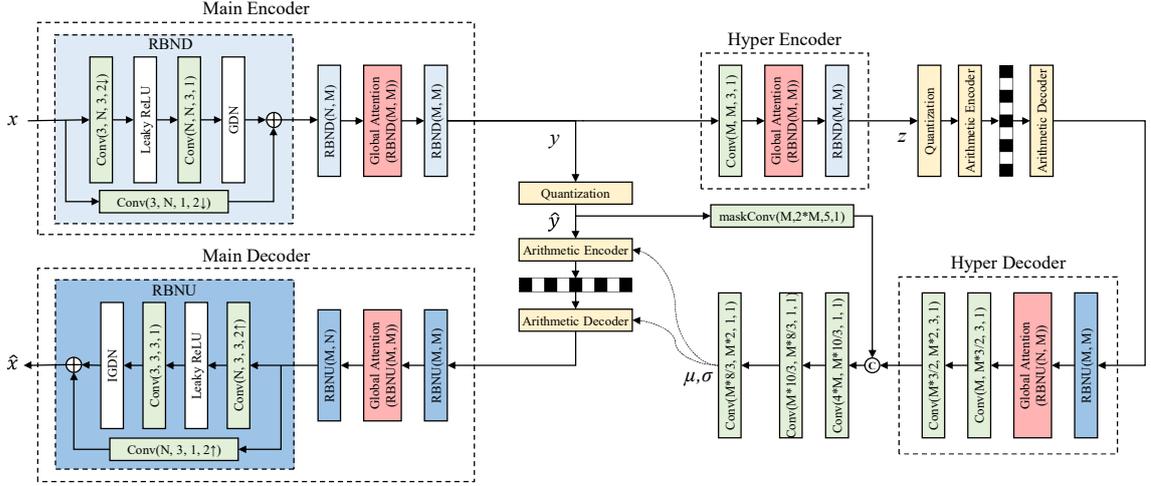

**Fig. 7.** The architecture of our proposed network with global attention module. *Conv* (*inch, outch, kernelsize, stride*) is a convolutional layer with the kernel size of *kernelsize × kernelsize* and a stride of *stride*. *Global Attention* (*conventionalnetwork*) represents the global attention module using *conventionalnetwork* as resampling network structure.

## V. PROPOSED END-TO-END COMPRESSION NETWORK

Based on the proposed data preprocessing scheme and focused plenoptic image dataset, our end-to-end network accompanied by the global attention module is proposed.

### A. Network Architecture

As depicted in Fig. 7, our network adopts a hierarchical hyperprior-context network [36] as the backbone, where the main encoder adopts 4 down-sampling operations mapping the given input $x$ to the latent representation $y$. An auxiliary hyperprior autoencoder is used to capture the spatial corrections among $y$ into side information $z$. Up-sampled quantized $\hat{z}$ combined with a context model $Cm$ is utilized to estimate the parameters of Gaussian distribution of each element of quantized latent representation $\hat{y}$ for lossless entropy compression. Then the main decoder reconstructs the image $\hat{x}$ by up-sampling $\hat{y}$. Following [36], residual bottleneck (RBN) structure with efficient receptive field is modified as RBN down-sampling (RBND) with stride convolution and RBN up-sampling (RBNU) with sub-pixel convolution in the encoder and decoder, respectively, to improve rate-distortion (RD) performance. Generalized divisive normalization (GDN) [34] and its inverse format IGDN, are used as the local normalization to remove statistical dependencies in RBN.

Considering the long-distance and pixel-wise redundancy among the microimages in plenoptic images, the third and first down-sampling/up-sampling networks in the main and hyperprior autoencoder are modified respectively with the proposed global attention module to extract the compact representations for better compression efficiency, as shown in Fig. 7.

Learned image compression aims to minimize the bitrate consumption and expected distortion $\hat{x}$ respect to $x$. Therefore, the loss function is given by:

$$\mathcal{L} = \mathcal{R}(y) + \mathcal{R}(z) + \lambda \cdot \mathcal{D}(x, \hat{x}), \quad (3)$$

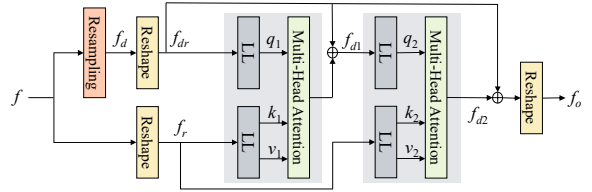

**Fig. 8.** Global attention module. The feature map $f$ and resampled feature map $f_d$ are reshaped into pixels-wise vector $f_r$ and $f_{dr}$. The relevant latent vectors of $f_{d1}$ are fetched with the multi-head global attention with the query $q_1$ derived from $f_{dr}$ and the key $k_1$, the value $v_1$ derived from $f_r$ by linear layer. With 2 times global attention computing, resampling result $f_o$ is obtained by reshaping output of attention $f_{d2}$ back to multi-dimension. The global attention module is beneficial to capture the global information in down-sampling/up-sampling operations.

where $\lambda$ is the Lagrange multiplier to control the trade-off between bitrate and distortion; $\mathcal{R}(\cdot)$ represents the bitrate; and $\mathcal{D}(x, \hat{x})$ is the distortion between input image $x$ and reconstructed image $\hat{x}$.

### B. Global Attention Module

Due to the noticeable long-distance and pixel-wise redundancy among the microimages from the imaging principle, the key point of improving focused plenoptic compression efficiency is to extract the compact feature map with a large receptive field. And for learned compression, the down-sampling operations generally cause information loss affecting the reconstruction quality.

To address this problem, we propose the global attention module with pixel-wise vector correlation to explore the global dependencies among input feature maps $f \in \mathbb{R}^{C \times H \times W}$ in the down-sampling/up-sampling process, where *C, H*, and *W* are the number of channels, the height, and the width, respectively. Since down-sampling process in the encoder and up-sampling extraction in the decoder are symmetric in the end-to-end network, we illustrate the global attention module in the down-sampling manner without loss of generality.

The proposed global attention module is depicted in Fig. 8. First, down-sampled feature map $f_d \in \mathbb{R}^{C \times \frac{H}{2} \times \frac{W}{2}}$ is generated from $f$ by a convolution layer with stride of 2 or other down-sampling network structures. Then, feature map $f$ and $f_d$ apply a regular reshaping strategy which merges the dimensions of height and width and permutes the channels dimension with merged height-width dimension to transform into pixel-wise vectors. And the global attention computing is utilized for explore the correlation between the reshaped feature maps $f_r \in \mathbb{R}^{HW \times C}$ and $f_{dr} \in \mathbb{R}^{\frac{HW}{4} \times C}$, The global attention computing can be expressed as:

$$q_1 = LL(f_{dr})$$
$$k_1, v_1 = LL(f_r) \qquad (4)$$
$$f_{d1} = Softmax(\frac{q_1 k_1^T}{\sqrt{C}})v_1 + f_{dr},$$

where the query $q_1 \in \mathbb{R}^{\frac{HW}{4} \times C}$ is derived from $f_{dr}$ through the linear layer (LL); the key $k_1 \in \mathbb{R}^{HW \times C}$ and the value $v_1 \in \mathbb{R}^{HW \times C}$ are acquired from $f_r$. The global attention computing maps the query and the set of key-value pairs to the output in pixel-wise vector format. Thus, the global attention computing enables to generate the compact output $f_{d1}$ by drawing the global dependencies of input $f_r$ with a skip connection of $f_{dr}$. The capacity of attention is enhanced by the multi-head attention computing [61]. With 2 times global attention computing, the final output of global attention $f_{d2} \in \mathbb{R}^{\frac{HW}{4} \times C}$ adding with $f_{dr}$ is reshaped back to the high dimensional feature map $f_o \in \mathbb{R}^{C \times \frac{H}{2} \times \frac{W}{2}}$. Thus, the long-distance content information among the feature map $f$ is represented compactly in feature map $f_o$ by the large receptive attention module to affect RD performance.

The global attention module uses RBND and RBNU as the resampling component in the encoder and decoder. Considering the data size of the proposed image dataset and the complexity of attention module, we only insert the global attention module in the third down-sampling/up-sampling stage of the main autoencoder and first down-sampling/up-sampling stage of the hyper autoencoder, as shown in Fig. 7.

## VI. EXPERIMENTS AND ANALYSIS

In this section, the effectiveness of the proposed method is demonstrated by comparing with SOTA hybrid codecs and learned models. First, the implementation details and test conditions are described. Then, the rate-distortion results and qualitative results are provided and analyzed. Thirdly, ablation study is conducted in terms of the effectiveness of the global attention module.

*A. Implementation Details*

**Training:** For training, we use 1877 preprocessed plenoptic images in "FPI2k" from all categories, and sequentially crop them into 75080 patches with the size of 384 × 384 × 3. Each patch includes 8 × 8 complete cropped microimages. The number of channels $N$ and $M$ are set to 128 and 192 in Fig. 7, respectively. Our models are optimized using MSE as the distortion metric in Eq. (3). We trained 6 models in total with λ = (0.1, 0.05, 0.025, 0.01, 0.005, 0.001). All the models are trained for 50 epochs using the Adam with a batch size of 4. The initial learning rate is set to $1 \times 10^{-4}$ and dynamically reduced by ReduceLROnPlateau with the patience of 5.

**Evaluation:** For comparison, 2 plenoptic images, the first frames of *Cars* [59] and *Matryoshka* [62] in MPEG-I Visual test video sequences, and 18 images in "FPI2k" are selected as the test images. The test set does not overlap with the training set. In inference, the preprocessed plenoptic images are padded to the multiple of 384 with zero and cropped into 384 × 384 patches in learned compression for computational efficiency. 5 × 5 sub-aperture images are rendered from each reconstructed plenoptic image via the TSPC rendering tool [39]. All the central sub-aperture images of test images are shown in Fig. 9.

*B. Test Conditions*

To evaluate the effectiveness of the proposed method and data preprocessing scheme, we compare our method with hybrid intra-coding tools, learned image compression models, and sub-aperture image compression methods. Reference software HM-16.22 [63], HM-16.9_SCM8.0 [64], and VTM-10.0 [65] are selected as the intra coding tools on original and preprocessed plenoptic image compression for general hybrid standards: H.265/HEVC [52], HEVC Screen Content Coding Extension (SCC) [46], and H.266/VVC [53], respectively. The macropixel-based method co-located single-block prediction (CSP) mode [16] on HEVC SCC for traditional plenoptic image are chosen for original and preprocessed plenoptic image compression. The imaging-principle-based method [18] based on HEVC and zoomed imaging-principle-guided prediction using SCC profile [17] for focused plenoptic images are utilized for plenoptic image compression. Four learned compression methods, including the basic factorized network Ballé *et al*. [34], the hyperprior network Ballé *et al*. [35], joint context-hyperprior work Minnen *et al*. [37], and Cheng *et al*. [36] which is the first learned work surpassing VVC intra-coding, are selected for preprocessed plenoptic image compression comparison. All the learned image compression models adopt the same training and evaluation strategy as our method. The sub-aperture image reordering image compression of spiral scan order [27] and raster scan order [26] are compared on HM-16.22. In total, 17 testing cases are defined and tested, as listed in Table II. The resolution of plenoptic images, preprocessed plenoptic images and sub-aperture images are 4080 × 3068, 3168 × 2016, and 1149 × 830 × 5 × 5, respectively.

For a fair comparison, the common test conditions [5] in MPEG-I lenslet video coding is used to evaluate the effectiveness of testing cases for intra coding, learned image compression, and sub-aperture image compression. As depicted in Fig. 10, the distortions are measured with sub-aperture images rendered from original plenoptic images in RGB color space. The distortion between the reference sub-aperture images and the reconstructed sub-aperture images is measured by the average of PSNR or MS-SSIM of each sub-aperture image. The distortion calculation is given by:

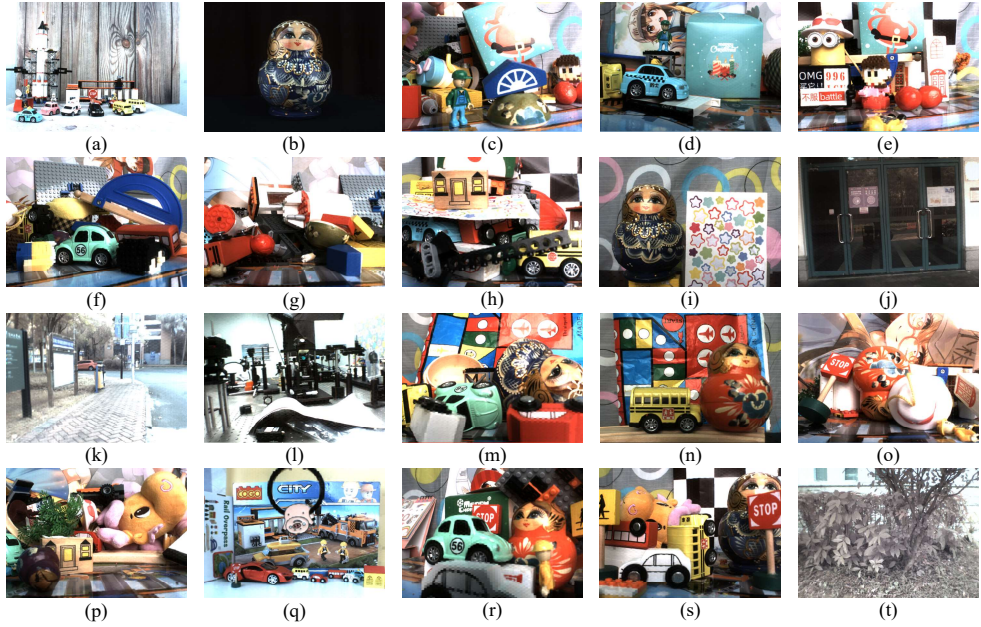

**Fig. 9.** Central views of 20 test images. (a) I01, and (b) I02, are the first frame of MPEG I-Visual test video sequences "Cars" [59] and "Matryoshka" [62], respectively; (c)-(u) are images from "FPI2k", which are (c) I03, Box-1; (d) I04, Box_7; (e) I05, Box-28; (f) I06, Building_block_5; (g) I07, Building_block_86; (h) I08, Building_block_141; (i) I09, Calendar_9; (j) I10, Door_8; (k) I11, Indicator_6; (l) I12, Laboratory_65; (m) I13, Matryoshka_93; (n) I14, Matryoshka_169; (o) I15, Matryoshka_170; (p) I16, Plush toy_135; (q) I17, Toy car_1; (r) I18, Toy car_197; (s) I19, Toy car_502; (t) I20, Tree_29.

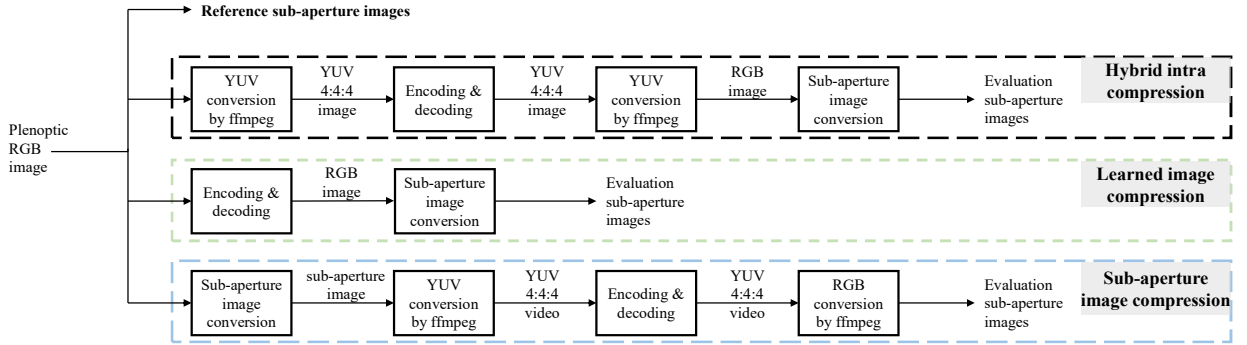

**Fig. 10.** Test flow chart for focused plenoptic image compression performance comparison following the common test conditions in MPEG lenslet video coding [5]. The sub-aperture images rendered from original plenoptic images are set as the reference to compute average distortion. The conversion and compression flow of hybrid intra codecs, learned plenoptic image compression and sub-aperture image compression methods are circled by black-dashed rectangle, green-dashed rectangle and blue-dashed rectangle, respectively.

$$PSNR = \frac{1}{5 \times 5} \sum_{k=1}^{5} \sum_{j=1}^{5} PSNR(i,j)$$
$$MS\text{-}SSIM = \frac{1}{5 \times 5} \sum_{k=1}^{5} \sum_{j=1}^{5} MS\text{-}SSIM(i,j) \quad , \quad (4)$$

where $PSNR(i,j)$ / $MS\text{-}SSIM(i,j)$ is the objective PSNR/MS-SSIM metric computed for the $(i,j)^{th}$ sub-aperture image. The RD performance is measured in terms of BD-rate [66]. The bitrate is defined by bit-per-pixel (bpp) which is calculated via dividing the number of bits by the total number of pixels of the original plenoptic image.

C. *Experimental Results*

*1) Rate-distortion Performance:* The compression efficiency of the proposed data preprocessing scheme and network are evaluated objectively in this subsection. First, the efficiency comparison of the original and preprocessed plenoptic images on different hybrid standards is listed in Table III. It can be found that the preprocessed images can improve compression efficiency significantly. As shown in the table, the data preprocessing scheme can achieve 40.24%, 34.58%, 43.96%, and 35.95% bitrate reduction on average for HEVC, CSP mode, SCC, and VVC intra coding, respectively, because preprocessed plenoptic images discard 48.98% sub-aperture image ineffective pixels from original plenoptic images. Besides, the HEVC intra coding on preprocessed images shows superior compression efficiency compared to VVC intra coding on original images with an average of 17.47% bitrate reduction.

TABLE II
TESTING CASES AND CODING CONFIGURATION

| Input format | Testing cases | Configuration |
|---|---|---|
| Sub-aperture images | *HEVC+SOP* | "HEVC" + raster scan order proposed in [26] with Low Delay profile |
| | *HEVC+SPR* | "HEVC" + spiral scan order proposed in [27] with Low Delay profile |
| Original plenoptic images (OPI) | *HEVC+OPI* | "HEVC" RExt intra profile [63] |
| | *CSP+OPI* | method CSP proposed in [16] |
| | *IIP* | method IIP proposed in [18] |
| | *ZIIP* | method + SCC profile proposed in [17] |
| | *SCC+OPI* | "HEVC-SCC" intra profile [64] |
| | *VVC+OPI* | "VVC" intra profile [65] |
| Preprocessed plenoptic images (PPI) | *HEVC+PPI* | "HEVC" RExt intra profile [63] |
| | *CSP+PPI* | method proposed in [16] |
| | *SCC+PPI* | "HEVC-SCC" intra profile [64] |
| | *VVC+PPI* | "VVC" intra profile [65] |
| | *Factor* | method proposed in [34] |
| | *Hyper* | method proposed in [35] |
| | *Joint* | method proposed in [37] |
| | *Attention* | method proposed in [36] |
| | *Proposed* | proposed network |

Secondly, the compression efficiency of the proposed network combined with the data preprocessing scheme is evaluated in Table IV. As shown in the table, the proposed method can save 76.92% and 74.37% bitrate on average compared to SOTA focused plenoptic image compression methods *IIP* and *ZIIP*, respectively. And, the proposed method achieves an average bitrate reduction of 62.57%, 62.92%, 53.28%, and 51.67% bitrate saving compared to HEVC, CSP, SCC, and VVC intra coding on preprocessed plenoptic images, respectively. Furthermore, it outperforms Ballé *et al.* [34], Ballé *et al.* [35], Minnen *et al.* [37], and Cheng *et al.* [36] by 62.66%, 52.44%, 36.54%, and 18.73% bitrate reduction, respectively. The efficiency improvement between the proposed method and *Attention* reveals the proposed global attention module with large receptive field is more beneficial than the local attention mechanism to the plenoptic image compression. It can be found that the proposed method performs worse than *HEVC+SPR* on test image I01, I02, I10, and I20 in Table IV. The reason for the inferior performance lies in two aspects. One reason is that objects and the camera are far from each other resulting in a small disparity among the views and the scenes in I01, I02, I10, and I20 are captured with much less texture which is beneficial to the conventional inter-prediction architecture that exploits interview correlation among the reordered sub-aperture images.

TABLE III
BD-RATE COMPARISON AMONG THE ORIGINAL AND PREPROCESSED PLENOPTIC IMAGE COMPRESSION METHODS

| Images | HEVC+PPI vs. HEVC+OPI | CSP+PPI vs. CSP+OPI | SCC+PPI vs. SCC+OPI | VVC+PPI vs. VVC+OPI | HEVC+PPI vs. VVC+OPI |
|---|---|---|---|---|---|
| I01 | -40.3% | -34.0% | -42.6% | -37.6% | -22.0% |
| I02 | -39.9% | -37.1% | -45.4% | -37.9% | -20.2% |
| I03 | -43.0% | -37.3% | -45.7% | -38.3% | -20.3% |
| I04 | -39.0% | -30.8% | -41.2% | -34.0% | -12.4% |
| I05 | -43.8% | -31.8% | -46.9% | -39.3% | -20.4% |
| I06 | -40.0% | -34.7% | -42.9% | -34.0% | -14.6% |
| I07 | -45.3% | -38.7% | -46.9% | -40.6% | -23.1% |
| I08 | -44.5% | -39.9% | -46.3% | -39.9% | -21.6% |
| I09 | -38.3% | -31.7% | -42.7% | -33.8% | -14.6% |
| I10 | -38.8% | -33.1% | -42.8% | -35.3% | -16.7% |
| I11 | -35.8% | -29.9% | -42.0% | -31.7% | -16.7% |
| I12 | -41.5% | -37.2% | -46.9% | -38.4% | -22.4% |
| I13 | -41.4% | -35.7% | -44.9% | -35.9% | -16.2% |
| I14 | -39.7% | -34.5% | -42.1% | -34.7% | -13.8% |
| I15 | -42.9% | -37.6% | -45.7% | -38.4% | -18.3% |
| I16 | -37.6% | -32.2% | -41.1% | -33.2% | -15.3% |
| I17 | -36.0% | -30.4% | -42.6% | -31.9% | -10.6% |
| I18 | -38.4% | -32.6% | -42.3% | -33.0% | -11.6% |
| I19 | -42.3% | -36.9% | -44.2% | -37.2% | -18.1% |
| I20 | -36.2% | -35.5% | -43.9% | -33.9% | -20.4% |
| **Avg.** | **-40.24%** | **-34.58%** | **-43.96%** | **-35.95%** | **-17.47%** |

It can be proved that the proposed method can achieve 66.5% bitrate reduction compared to *HEVC+SPR* on I18 taken from a close shooting distance while our method achieves 56.0% bitrate increment compared to *HEVC+SPR* on I10 taken from outdoors. The other reason is that preprocessed plenoptic images contain more information ($d = 48 > d_{\min} = 40$) with a total 44% pixel increment than the internal regions of cropped microimages to render images.

TABLE IV
BD-RATE COMPARISON FOR PROPOSED METHOD

| Images | *Proposed* | | | | | | | | | | | |
|---|---|---|---|---|---|---|---|---|---|---|---|---|
| | vs. IIP | vs. ZIIP | vs. HEVC+PPI | vs. CSP+PPI | vs. SCC+PPI | vs. VVC+PPI | vs. Factor | vs. Hyper | vs. Joint | vs. Attention | vs. HEVC+SOP | vs. HEVC+SPR |
| I01 | -65.1% | -60.6% | -45.5% | -45.9% | -32.1% | -31.9% | -50.3% | -35.1% | -22.8% | -10.9% | 20.2% | 38.1% |
| I02 | -65.1% | -63.8% | -44.3% | -44.4% | -35.9% | -29.4% | -72.3% | -54.6% | -28.9% | -11.8% | 22.9% | 36.8% |
| I03 | -81.5% | -79.4% | -67.6% | -68.0% | -60.5% | -57.6% | -62.5% | -54.8% | -39.3% | -18.6% | -57.5% | -51.9% |
| I04 | -78.1% | -74.7% | -66.1% | -66.5% | -55.3% | -55.3% | -64.2% | -53.9% | -36.1% | -20.0% | -27.7% | -16.7% |
| I05 | -81.0% | -78.3% | -66.0% | -66.4% | -57.5% | -55.2% | -64.3% | -55.7% | -41.2% | -18.9% | -54.2% | -47.7% |
| I06 | -81.1% | -78.0% | -68.2% | -68.6% | -59.1% | -58.3% | -65.9% | -57.1% | -42.0% | -21.2% | -51.1% | -43.8% |
| I07 | -82.4% | -79.4% | -68.5% | -68.8% | -59.1% | -57.8% | -65.7% | -56.3% | -40.1% | -18.1% | -50.0% | -43.0% |
| I08 | -80.1% | -78.2% | -64.3% | -64.7% | -57.8% | -53.5% | -63.1% | -53.6% | -38.0% | -14.5% | -68.9% | -64.9% |
| I09 | -79.1% | -76.9% | -66.9% | -67.3% | -58.2% | -57.5% | -61.2% | -53.1% | -37.9% | -22.6% | -24.6% | -13.7% |
| I10 | -67.9% | -66.6% | -54.8% | -55.3% | -43.6% | -41.2% | -56.9% | -46.4% | -32.3% | -20.6% | 29.3% | 56.0% |
| I11 | -74.3% | -72.3% | -63.4% | -63.6% | -52.0% | -53.9% | -57.4% | -49.5% | -35.4% | -20.6% | -14.3% | -5.3% |
| I12 | -75.0% | -72.4% | -55.4% | -55.8% | -46.2% | -43.8% | -56.7% | -45.3% | -27.5% | -13.5% | -40.0% | -29.7% |
| I13 | -80.8% | -78.2% | -67.3% | -67.7% | -59.2% | -57.1% | -65.3% | -56.7% | -41.4% | -16.2% | -61.2% | -55.1% |
| I14 | -87.1% | -85.5% | -78.3% | -78.6% | -73.2% | -71.5% | -62.8% | -55.0% | -38.7% | -18.2% | -42.1% | -34.2% |
| I15 | -81.3% | -79.2% | -67.4% | -67.9% | -60.0% | -56.3% | -75.7% | -69.3% | -54.9% | -36.1% | -67.1% | -61.2% |
| I16 | -77.1% | -75.0% | -63.6% | -63.9% | -55.7% | -54.0% | -60.1% | -51.2% | -35.1% | -17.2% | -36.5% | -27.6% |
| I17 | -72.1% | -66.6% | -57.5% | -57.8% | -40.8% | -43.8% | -61.5% | -47.7% | -33.8% | -16.8% | -22.6% | -12.4% |
| I18 | -78.3% | -76.4% | -65.7% | -66.2% | -57.9% | -55.0% | -67.7% | -56.0% | -38.9% | -17.7% | -70.9% | -66.5% |
| I19 | -79.6% | -77.6% | -65.5% | -65.8% | -58.6% | -55.3% | -63.4% | -54.6% | -36.8% | -20.0% | -58.0% | -51.9% |
| I20 | -71.3% | -68.3% | -55.0% | -55.2% | -42.9% | -45.0% | -56.1% | -42.9% | -29.7% | -21.1% | -5.6% | 6.5% |
| **Avg.** | **-76.92%** | **-74.37%** | **-62.57%** | **-62.92%** | **-53.28%** | **-51.67%** | **-62.66%** | **-52.44%** | **-36.54%** | **-18.73%** | **-34.00%** | **-24.41%** |

The RD curves of I01 "Cars" and I15 "Matryoshka_170" of 6 original plenoptic image compression testing cases including *HEVC+OPI*, *IIP*, *ZIPP*, *CSP+OPI*, *SCC+OPI*, and *VVC+OPI* are depicted in Fig. 11 accompanying with the proposed network. Although *VVC+OPI* achieves the best compression efficiency on 2 test images compared to the plenoptic image compression methods in terms of PSNR and MS-SSIM, the big performance gap between the proposed method and *VVC+OPI* demonstrates the superiority of the proposed method. To the best of our knowledge, our work is the first learned work to achieve better performance than the VVC intra compression on focused plenoptic image compression.

The compression efficiency among the 4 hybrid testing cases and 5 learned methods on preprocessed plenoptic images is shown in Fig. 12. The proposed method achieves the best performance compared to the hybrid codecs and learned image compression methods both on the metric of PSNR and MS-SSIM.

Two pseudo-video-based methods using zigzag and spiral scan order on HEVC are also compared, as shown in Fig. 13. The proposed method exhibits inferior compression performance at low bitrates (rate<0.2bpp) measured by PSNR and MS-SSIM, however, it manifests superior compression performance at high bitrates (rate>0.2bpp) of PSNR on I01 "Cars". Although the proposed method does not save bitrate when contrasted with reordering methods on I01 "Cars" in Table IV, it demonstrates promising compression performance at high test bitrates. On I15 "Matryoshka_170", the proposed method surpasses HEVC reordering methods concerning PSNR and MS-SSIM at all test bitrates.

*2) Subjective Comparison Results:* To demonstrate our method can generate visually pleasant results, the central views of reconstructed I01 of 9 testing cases with approximately 0.06 bpp and a compression ratio of 400:1 are demonstrated in Fig. 14. Although the proposed network is optimized by MSE in Eq. (3) which prefers to produce smooth reconstructions, our method achieves comparable results with *HEVC+SOP*, and outperforms *HEVC+OPI*, *VVC+OPI*, *CSP+PPI*, *VVC+PPI*, *Factor*, and *Attention*. The stop sign marked in green rectangle generated by our method appears much more natural than the other codecs. Besides, more details are preserved by our method for the texture regions, as shown in the enlarged background regions in Fig. 14.

*3) Execution Time Comparison Results:* To evaluate the computational complexity of the proposed technique, a comparison with the traditional codecs *HEVC+PPI*, *VVC+PPI*, and *HEVC+SPR* is conducted in terms of encoding and decoding time. All the testing cases are performed on the same machine equipped with Intel(R) Core(TM) i7-9700 CPU @ 3.00 GHz, 16GB RAM, and 64-bit Windows 10 operating system. The proposed method employs the same and isomorphic encoder and decoder which yields similar encoding and decoding time for different fixed-rate models. However, the encoding and decoding time differ when using different quantization parameters in traditional codecs. Four bitrates, labeled as $r_1$, $r_2$, $r_3$, and $r_4$ from low to high bitrate, are used for execution time comparison. The quantization parameters used for *HEVC+PPI* and *VVC+PPI* are 42, 37, 32, and 27; for *HEVC+SPR*, they are 31, 28, 25, and 22 to match the target bitrates. The proposed method uses fixed-rate models with Lagrange multiplier values of 0.005, 0.025, 0.05, and 0.1. The average execution time of each testing case and relative ratios between the proposed method and other testing cases under four bitrates on 20 test images are summarized in Table V. As listed in the table, although the encoding complexity of the proposed method presents an average of 2.7 times higher than that of *HEVC+PPI*, more than 60% of bitrate reduction can be achieved, as listed in Table IV. Its encoding complexity is 2.22 times and 1.51 times lower than that of *VVC+PPI* and *HEVC+SPR*, respectively, with much higher compression efficiency, which is very beneficial to applications. Currently, its decoding complexity is much higher than that of *HEVC+PPI*, *VVC+PPI*, and *HEVC+SPR*, which is mainly caused by the strict serial entropy coding modeling. This problem may be solved by applying advanced parallelization-friendly entropy coding methods [49, 67], which will be investigated further.

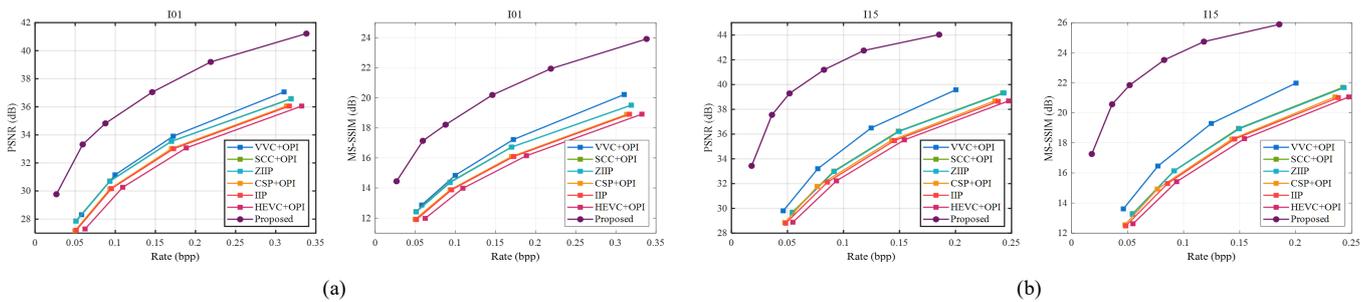

**Fig. 11.** RD performance evaluation of proposed network and original plenoptic image compression testing cases on (a) I01 "Cars" and (b) I15 "Matryoshka_170".

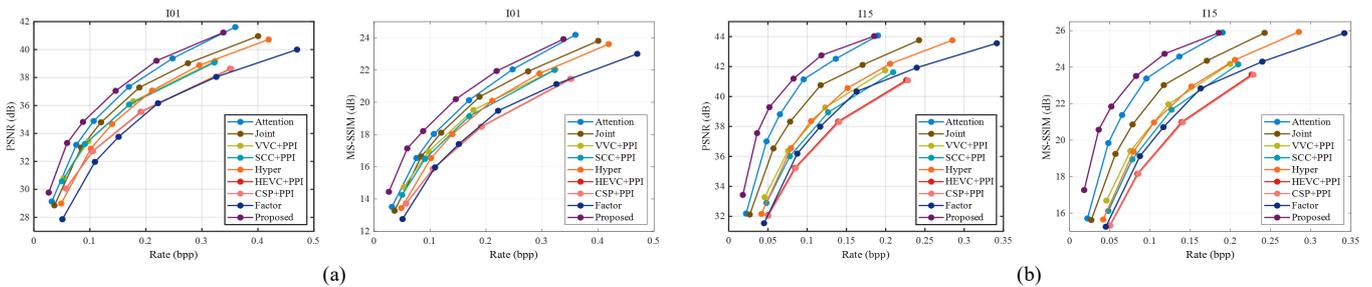

**Fig. 12.** RD performance evaluation of proposed network and preprocessed plenoptic image compression testing cases on (a) I01 "Cars" and (b) I15 "Matryoshka_170".

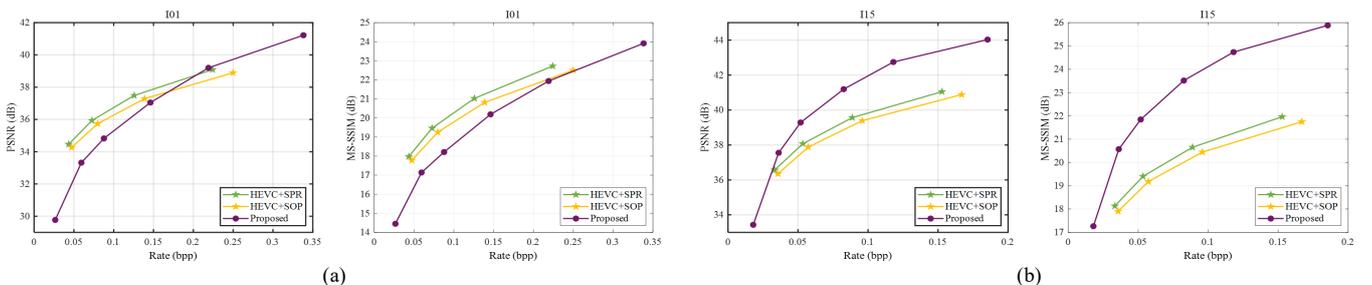

**Fig. 13.** RD performance evaluation of proposed network and sub-aperture-image-based compression testing cases on (a) I01 "Cars" and (b) I15 "Matryoshka_170".

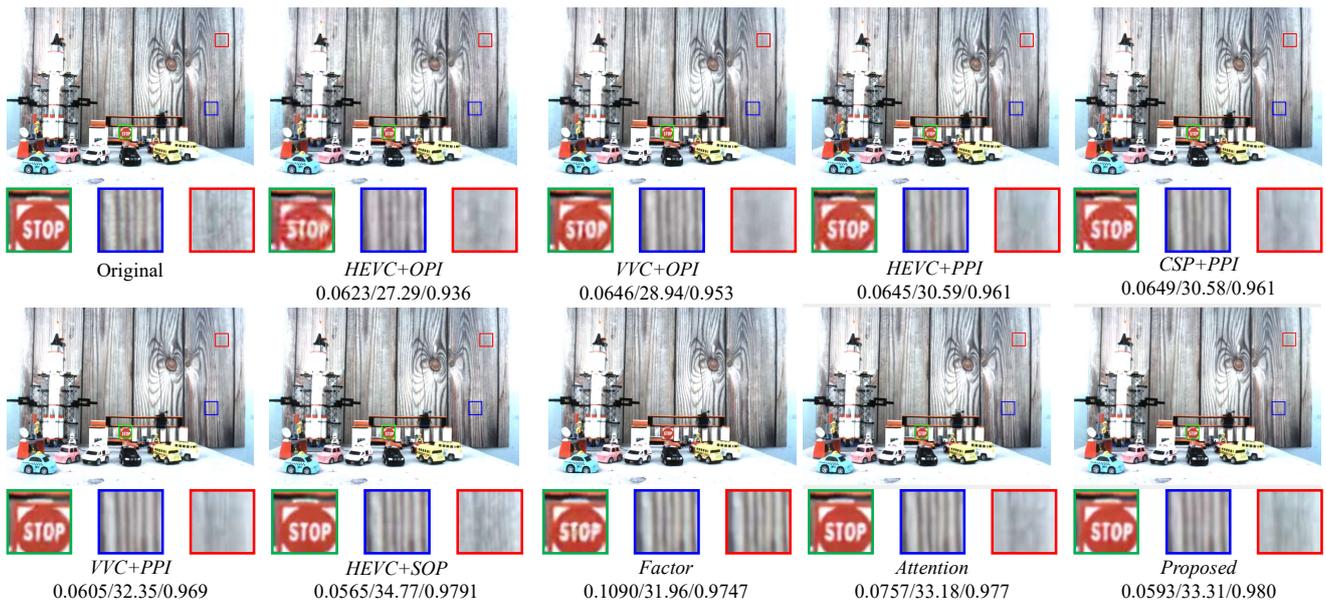

**Fig. 14.** Visualization of the central view of reconstructed plenoptic image I01 "Cars". Three selected areas are marked by green, blue and red rectangles respectively. The metrics are [bpp ↓ /PSNR ↑ /MS-SSIM ↑ ].

TABLE V
EXECUTION TIME AND RELATIVE RATIOS OF DIFFERENT CODING METHODS AT FOUR RATES

|  | Rate | Execution time | | | | Execution time ratio | | |
|---|---|---|---|---|---|---|---|---|
|  |  | HEVC+PPI | VVC+PPI | HEVC+SPR | Proposed | Proposed vs. HEVC+PPI | Proposed vs. VVC+PPI | Proposed vs. HEVC+SPR |
| Encoding Time | $r_1$ | 87.45s | 306.09s | 324.21s | 273.33s | 3.12 | 0.89 | 0.84 |
|  | $r_2$ | 92.42s | 574.52s | 368.82s | 252.32s | 2.72 | 0.44 | 0.68 |
|  | $r_3$ | 99.02s | 950.51s | 428.77s | 253.00s | 2.55 | 0.27 | 0.59 |
|  | $r_4$ | 107.89s | 1315.05s | 508.77s | 260.34s | 2.41 | 0.20 | 0.51 |
|  | **Avg.** | **96.70s** | **786.54s** | **407.64s** | **259.75s** | **2.70** | **0.45** | **0.66** |
| Decoding Time | $r_1$ | 0.33s | 0.52s | 0.80s | 372.57s | 1096.45 | 715.80 | 461.02 |
|  | $r_2$ | 0.38s | 0.61s | 0.89s | 339.29s | 884.73 | 552.28 | 378.09 |
|  | $r_3$ | 0.45s | 0.80s | 1.019s | 338.97s | 741.73 | 422.11 | 332.60 |
|  | $r_4$ | 0.49s | 0.98s | 1.15s | 354.37s | 713.30 | 361.48 | 305.89 |
|  | **Avg.** | **0.41s** | **0.72s** | **0.97s** | **351.30s** | **859.05** | **512.92** | **369.40** |

*D. Ablation Study*

To demonstrate the effectiveness of the proposed global attention module, we train and test the proposed network without global attention as baseline denoted as *Proposed w/o GA*. As shown in Fig. 15, the global attention module improves the compression efficiency by a large margin on the metrics of PSNR and MS-SSIM. Since the global attention module reduces bitrate consumption with similar distortion in the baseline, it proves that proposed module can capture the global information among feature maps to generate compact representation to improve compression efficiency. The superiority of global attention module lies in two core aspects. One is that the global attention module has a particularly large receptive field to capture the dependencies of the microimages, and the other is that pixel-wise vector attention can compute the correlation of the distinctive pixel distribution of object response in unfocused depths.

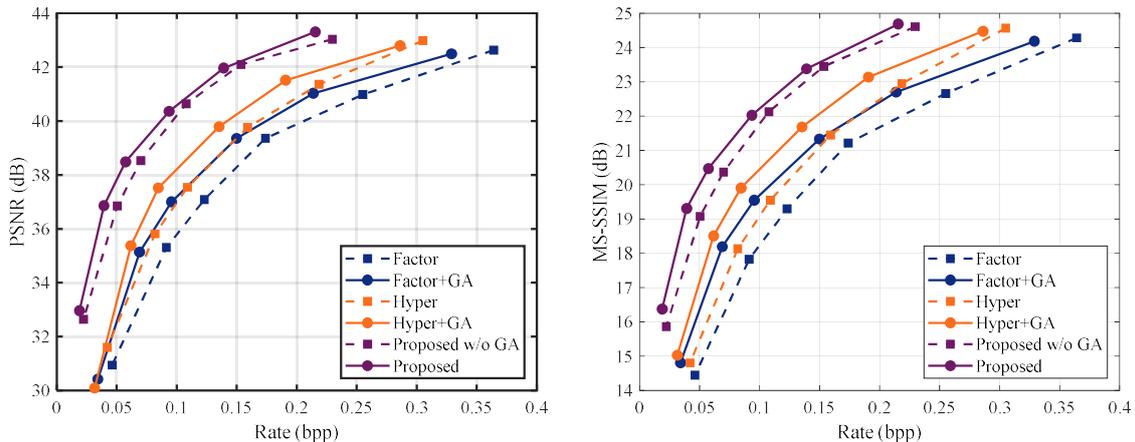

**Fig. 15.** Average RD performance evaluation of ablation study for global attention module on 20 test images.

To demonstrate the generality of the proposed module, we insert the global attention module in the third resampling structure of the main autoencoder and the first resampling operation of the hyper autoencoder in Ballé *et al.* [34] and Ballé *et al.* [35] similar to the proposed network. The modified networks are denoted as *Factor+GA* and *Hyper+GA* in Fig. 15. Compared to the baselines, the compression efficiency is also improved greatly for *Hyper+GA* and *Factor+GA* on test images in terms of PSNR and MS-SSIM, as shown in Fig. 15. And *Factor+GA* enhanced by the proposed global attention module achieves comparable results with *Hyper*. The performance improvement indicates that the global attention module can work as a plug-and-play component to enhance the existing models for plenoptic image compression.

## VII. CONCLUSIONS

In this paper, a lossy end-to-end network with the global attention module is proposed for focused plenoptic image efficient compression. First, a sub-aperture image lossless preprocessing scheme is designed according to the imaging principle to reshape the sub-aperture image effective pixels in each microimage and align the cropped microimages to the rectangular grid. Then, the proposed global attention module with large receptive field draws the global dependencies from the input feature maps in the resampling process to improve compression efficiency. Also, an image dataset with 1910 focused plenoptic images captured in real circumstances is built for model robust training and testing.

The proposed network achieves superior coding performance in compressing preprocessed plenoptic images in terms of the metric of PSNR and MS-SSIM compared to the existing SOTA coding standards including HEVC, HEVC SCC, and VVC with an average of 62.57%, 53.28%, and 51.67% bitrate reduction, respectively. And, it is the first learned work that shows better results than VVC intra coding on the focused plenoptic image compression. Our method outperforms the SOTA learned image compression model with 18.73% bitrate saving. Besides, the proposed method achieves comparable performance or even better with sub-aperture image compression on HEVC in terms of PSNR and MS-SSIM. The ablation study results reveal the effectiveness of the proposed global attention module which can work as a plug-and-play component to enhance the existing models for plenoptic image compression. For a different focused plenoptic camera, the values of its focal lengths of the main lens and microlens and the radius of each microlens are needed to change the EPA radius coefficient, cutting size in the proposed preprocessing, and the size of training patches containing the complete microimages for the network training and inference. Then, the network should be retrained to adapt network parameters to the variation in the intensity distribution. In the future, we will improve the size and diversity of the image dataset to promote the relative research fields. Also, we will explore other factors that utilize the distinctive distribution of focused plenoptic images to improve the compression efficiency, such as the local attention mechanism with a large receptive field with dilation convolution.

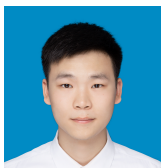

**Kedeng Tong** is currently pursuing the M.E. degree with Shenzhen International Graduate School, Tsinghua University. He has published several papers, e.g. ICASSP. His research interest is light-field compression, image compression.

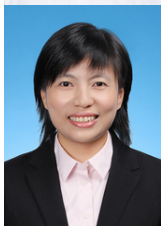

**Xin Jin** (S'03–M'09-SM'11) received the M.S. degree in communication and information system and the Ph.D. degree in information and communication engineering, both from Huazhong University of Science and Technology, Wuhan, China, in 2002 and 2005, respectively.

From 2006 to 2008, she was a Postdoctoral Fellow with The Chinese University of Hong Kong. From 2008 to 2012, she was a Visiting Lecturer with Waseda University, Fukuoka, Japan. Since Mar. 2012, she has been with Shenzhen International Graduate School, Tsinghua University, China, where she is currently a professor. Her current research interests include computational imaging and power-constrained video processing and compression. She has published over 170 conference and journal papers.

Dr. Jin is an IEEE Senior Member, and a member of SPIE and ACM. She is the Distinguished Professor of Pengcheng Scholar. She received Gold Medal of International Exhibition of Inventions of Geneva in 2022, the second prize of National Science and Technology Progress Award in 2016, the first prize of Guangdong science and technology award in 2015 and ISOCC Best Paper Award in 2010.

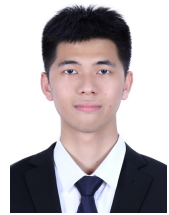

**Yuqing Yang** is currently pursuing the M.E. degree with Shenzhen International Graduate School, Tsinghua University. His research interest is video compression.

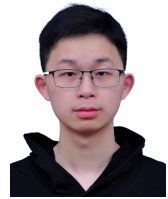

**Chen Wang** is currently pursuing the M.E. degree with Shenzhen International Graduate School, Tsinghua University. His research interest is video compression.

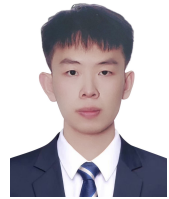

**Jinshi Kang** is currently pursuing the M.E. degree with Shenzhen International Graduate School, Tsinghua University. His research interest is light field defect inspection.

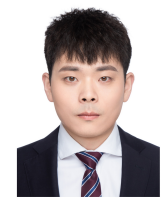

**Fan Jiang** received the M.E. degree with Shenzhen International Graduate School, Tsinghua University. He has published several papers, e.g. ICIP, VCIP, and T.BC. His research interest is light-field compression.